\documentclass[lettersize,journal]{IEEEtran}
\usepackage{amsmath,amsfonts}
\usepackage{algorithmic}
\usepackage{algorithm}
\usepackage{array}
\usepackage[caption=false,font=normalsize,labelfont=sf,textfont=sf]{subfig}
\usepackage{textcomp}
\usepackage{stfloats}
\usepackage{url}
\usepackage{verbatim}
\usepackage{graphicx}
\usepackage{cite}
\usepackage{multibib}
\usepackage{color}
\usepackage{multirow}
\usepackage{multicol}
\usepackage{float} 
\usepackage{booktabs}
\usepackage{arydshln}
\definecolor{Liam1}{RGB}{157,195,230} 
\definecolor{Liam2}{RGB}{146,208,080}
\definecolor{Liam3}{RGB}{247,169,136}
\definecolor{Liam4}{RGB}{255,230,153}

\begin{document}

\title{BLISS: Robust Sequence-to-Sequence Learning via\\ Self-Supervised Input Representation}

\author{Zheng Zhang$^\dagger$,
        Liang Ding$^\ddagger$,
        Dazhao Cheng$^\ddagger$,~\IEEEmembership{Senior Member,~IEEE}\\
        Xuebo Liu,
        Min Zhang,
        and~Dacheng Tao,~\IEEEmembership{Fellow,~IEEE}
\IEEEcompsocitemizethanks{
\IEEEcompsocthanksitem Z. Zhang and D. Cheng are with the Institute of Artificial Intelligence, School of Computer Science, Wuhan University, Wuhan, China (e-mail:zzhang3031@whu.edu.cn; dcheng@whu.edu.cn).
\IEEEcompsocthanksitem L. Ding and D. Tao are with the JD Explore Academy at JD.com, Beijing, China (e-mail: dingliang1@jd.com; dacheng.tao@jd.com).
\IEEEcompsocthanksitem X. Liu and M. Zhang are with the Institute of Computing and Intelligence, Harbin Institute of Technology, Shenzhen, China (e-mail:  liuxuebo@hit.edu.cn; zhangminmt@hotmail.com).

\IEEEcompsocthanksitem $^\dagger$Work was done when Zheng Zhang interning at JD Explore Academy; $^\ddagger$Corresponding Authors: Liang Ding (e-mail: dingliang1@jd.com) and Dazhao Cheng (e-mail: dcheng@whu.edu.cn).

}

}

\markboth{Journal of \LaTeX\ Class Files,~Vol.~14, No.~8, August~2021}%
{Shell \MakeLowercase{\textit{et al.}}: A Sample Article Using IEEEtran.cls for IEEE Journals}


\maketitle
\begin{abstract}
Data augmentations (DA) are the cores to achieving robust sequence-to-sequence learning on various natural language processing (NLP) tasks. However, most of the DA approaches force the decoder to make predictions conditioned on the perturbed input representation, underutilizing supervised information provided by perturbed input. In this work, we propose a framework-level robust sequence-to-sequence learning approach, named BLISS, via self-supervised input representation, which has the great potential to complement the data-level augmentation approaches. The 
key idea is to supervise the sequence-to-sequence framework with both the \textit{supervised} (``input$\rightarrow$output'') and \textit{self-supervised} (``perturbed input$\rightarrow$input'') information. We conduct comprehensive experiments to validate the effectiveness of BLISS on various tasks, including machine translation, grammatical error correction and text summarization. The results show that BLISS outperforms significantly the vanilla Transformer and consistently works well across tasks than the other five contrastive baselines. Extensive analyses reveal that BLISS learns robust representations and rich linguistic knowledge, confirming our claim. Source code will be released upon publication.



\end{abstract}


\section{Introduction}
Sequence-to-sequence learning~\cite{sutskever2014sequence} has advanced the state-of-the-art in various natural language processing (NLP) tasks, such as machine translation~\cite{bahdanau2014neural,wu2016google,Vaswani:17}, grammatical error correction~\cite{Kiyono:19,Kaneko:20} and text summarization~\cite{Wang:19,zhang2019pegasus}. 
Sequence-to-sequence learning models are generally implemented with an encoder-decoder framework, in which the encoder summarizes the input sentence, and predictions of the decoder are correspondingly supervised by matching the cross-entropy of ground truth. 
That is, the existing sequence-to-sequence learning frameworks are supervised by the  \textit{direct correlation between the input and the output}.

To achieve robust sequence-to-sequence learning, many data augmentation methods~\cite{Kobayashi:18,Wu:19,Gao:19,Cheng:20,Chen:21b,Morris:20} are proposed to enrich the training datasets by automatically or manually creating the perturbed input. For example, EDA~\cite{Wei:19} introduces some simple data augmentation strategies, e.g. insert, swap and deletion, working well for the low-resource settings. To avoid the generated data deviating from the original context, language models are employed to generate the substitutions for the subset of the input sentence~\cite{Kobayashi:18,Wu:19,Gao:19}. Besides, the adversarial techniques are also adopted to generate the adversarial samples to enhance the model generalization ~\cite{Cheng:20,Chen:21b,Morris:20}. 
Although those data-level approaches are straightforward and easy to use, all the above methods \textit{force the decoder to make lexical choices conditioned on the perturbed input representation}, which underutilizes supervised information provided by perturbed input.

In response to this problem, we propose a framework-level robust approach to make the most of the perturbed input in sequence-to-sequence learning via self-supervised input representation. The key idea is to supervise the sequence-to-sequence framework with both the \textit{transformation} from inputs to outputs, and the \textit{correlation} between the perturbed input and its original input. 
In particular, we employ two extremely simple and effective data augmentation techniques, i.e. shuffle and replacement, as the input perturbing function. 
Then, we propose a smoothness controller to harness the perturbing degree.
Based on the perturbed input, we correspondingly design a self-supervised mechanism upon the top of the encoder, where we choose the token prediction and position prediction as two objectives to restore the perturbed subset.
By doing so, we can achieve robust sequence-to-sequence learning by fully exploiting the supervised (``input$\rightarrow$output'') and self-supervised (``perturbed input$\rightarrow$input'') information.


We validated our approach on several sequence-to-sequence NLP tasks in Section~\ref{subsec:mainresults}, including machine translation~\cite{Bahdanau:15, Vaswani:17}, grammatical error correction~\cite{Wang:19, Zhang:20} and text summarization~\cite{Kiyono:19, Kaneko:20}, across five datasets. The experimental results show that our proposed BLISS significantly outperforms the vanilla Transformer and consistently works well across tasks than other five 
competitive baselines. Experiments on translation show that our proposed BLISS yields consistent improvements, ranging from 0.6 up to 0.8 BLEU points. As for correction and summarization tasks, we achieve +2.0 F$_{0.5}$ and +0.5 Rouge-L improvements against strong Transformer models, demonstrating the effectiveness and universality of our approach. 
In addition, we conducted comprehensive analyses in Section~\ref{subsec:analysis} to understand when and why our BLISS works. Furthermore, we showed that our framework-level self-supervised BLISS is complementary to some existing augmentation approach, e.g. SwitchOut~\cite{Wang:18}. Also, our BLISS is robust to inference noises and hyper-parameters compared to baselines. Importantly, through probing task~\cite{Conneau:18}, we found that our model could preserve significantly rich linguistic knowledge against vanilla Transformer.
Our main contributions can be summarized as: 
\begin{itemize}
    \item We introduce a robust sequence-to-sequence learning framework via self-supervised input representation, which has the potential to complement existing data augmentation approaches.
    \item Our approach provides a unified framework to make the most of existing supervised signals, i.e. correlation between input and output, and self-supervised signals, i.e. self-supervisions between perturbed input and original input. 
    \item We empirically validate the the effectiveness and universality on extensive experiments across tasks and datasets. 
\end{itemize}

\section{Related Work}
Our work is inspired by two lines of research: \romannumeral1) self-supervised learning and \romannumeral2) data augmentation.

\paragraph{Self-Supervised Learning} Self-supervision signals have been widely investigated in language model pretraining and unsupervised learning. BERT~\cite{Devlin:19} propose the mask language model, where they substitute a subset of tokens in the input sentence by a special symbol [MASK], and then predicts the missing tokens by the residual ones. MASS~\cite{Song:19} presents a sequence-to-sequence pre-training framework, which takes non-mask tokens as the encoder input and leverages masked tokens as the decoder input as well as the to-be-predicted target.
STRUCTBERT~\cite{Wang:20} extends BERT by leveraging the structural information: word-level ordering and sentence level ordering.
SpanBERT~\cite{Joshi:20} masks random contiguous spans rather than individual tokens and additionally introduces span-boundary objective. Different from these works that apply self-supervisions to the cost pre-train stage and fine-tune them on the down-stream tasks, we design the self-supervision objectives for input sentence to complement the existing MLE generation objectives to achieve further improvement.  

Similar to our work, there exists several works that combine self-supervisions with from-scratch sequence-to-sequence model training. JM-S2S~\cite{Guo:20b} introduce mask task to non-autoregressive translation model to fully exploit the undertrained encoder. Monolingual data is used by self-supervisions in multilingual translation~\cite{Siddhant:20}. Self-supervised and supervised learning are combined to optimize the machine translation models especially for the rich-resource settings~\cite{Cheng:21}. Different from these works, we propose a plug-and-play self-supervised input representation approach for general sequence-to-sequence tasks, which could be used to complement any data augmentation approaches and consistently enhance the model performance.
\begin{figure}[tb]
    \centering
    \includegraphics[width=0.39\textwidth]{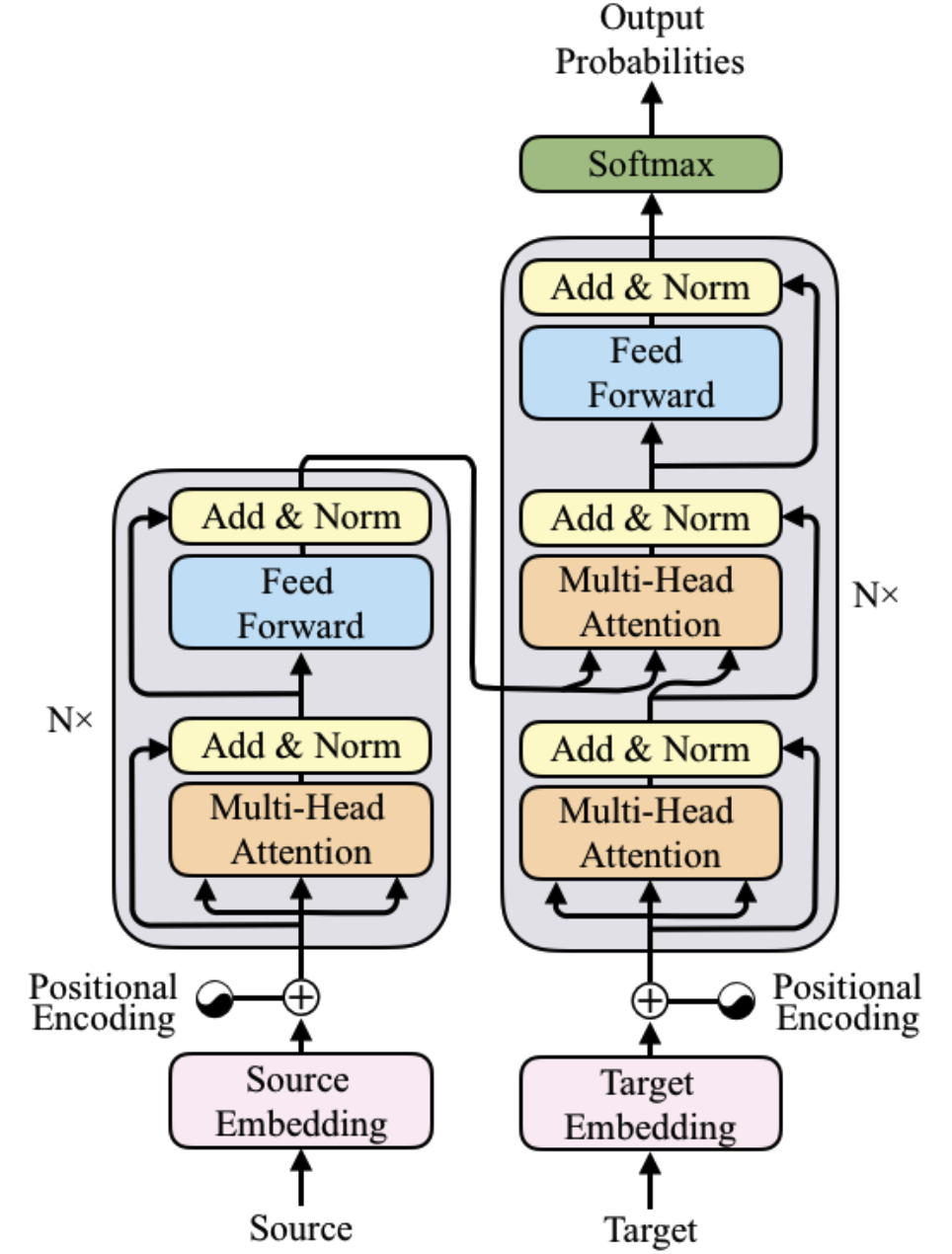}
    \caption{Illustration of the Transformer based encoder-decoder model.}
    \label{fig:transformer}
\end{figure}
\begin{figure*}[t] 
    \centering
    \includegraphics[width=0.8\textwidth,trim=0 50 0 80,clip]{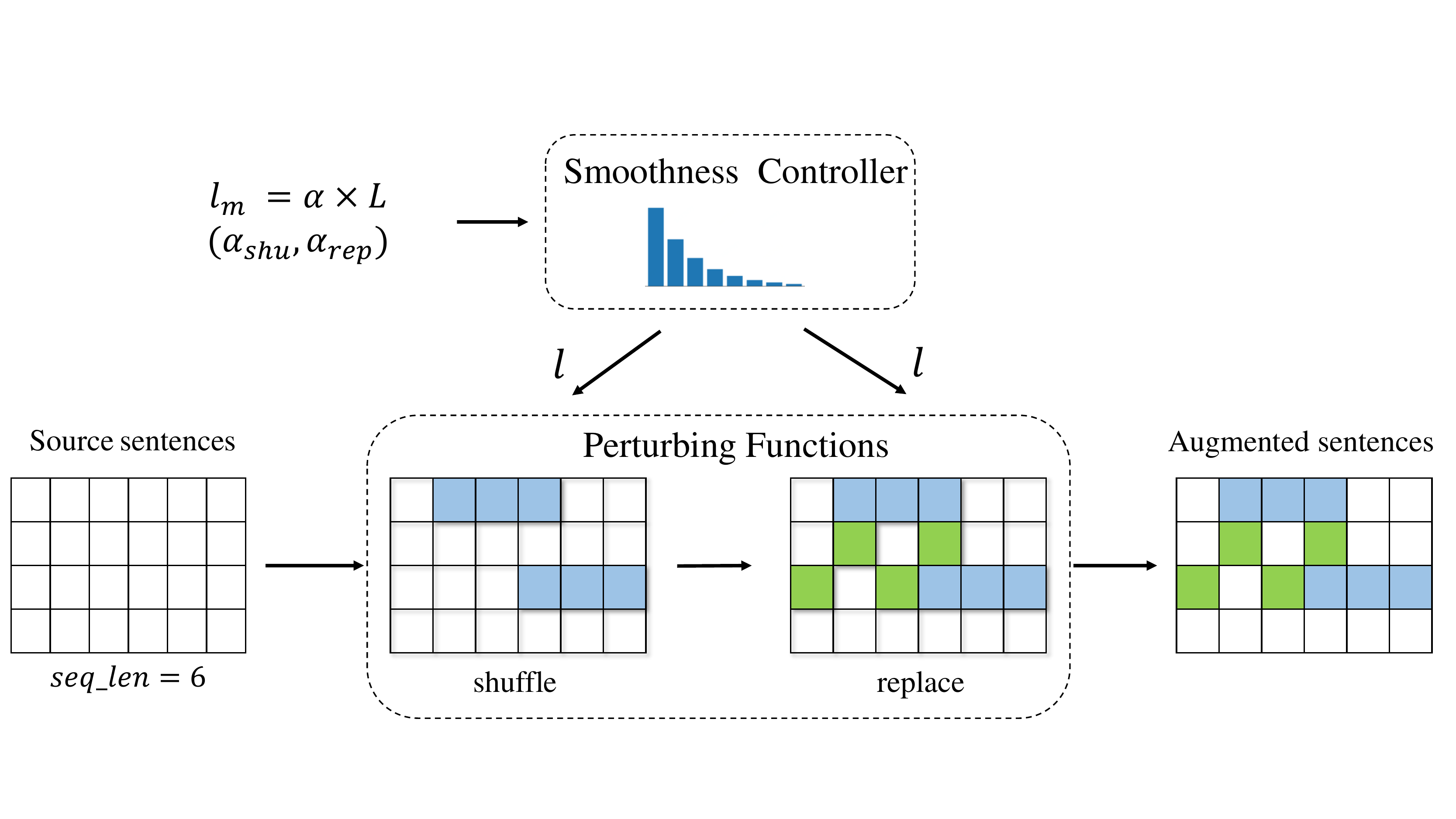} 
    \caption{Illustration of the proposed smooth augmented data generator in Section~\ref{subsec:smooth}, which consists of two components, i.e. perturbing functions and smoothness controller, represented by dashed rounded rectangles, respectively. The \textcolor{Liam1}{blue} block represent tokens been shuffled while the \textcolor{Liam2}{green} block represent tokens been replaced.} 
    \label{fig:smooth}
\end{figure*}

\paragraph{Data Augmentation} There exists some easy augmentation method, including randomly shuffling the words within a fixed window size to construct the perturbed sentence~\cite{Artetxe:18,Lample:18}, dropping some words randomly in the source sentence for learning an auto-encoder to help train the unsupervised NMT model~\cite{Iyyer:15}, replacing the word with a placeholder token or a word sampled from the frequency distribution of vocabulary~\cite{Xie:17}, mixing the token representations to encourage the diversified generation~\cite{Cao2021TowardsED} and other common NLP data scaling approaches~\cite{ding-etal-2021-improving,wang2022c3da}. These methods are usually useful in small datasets. However, some studies~\cite{Wei:19} experiments with easy data augmentation methods like randomly insert, swap and delete and they found that these simple methods take little effect with full datasets.
Besides, SwitchOut~\cite{Wang:18} replaces words in the source/target sentences with other words form the source/target vocabulary. 
SeqMix~\cite{Guo:20} mixs up pairs of source sentences or decoder input sentences. 
Our work significantly differs from these work. We do not predict the target lexicons conditioned on these perturbed input directly. Rather, we propose to recover the noised input with encoder, thus the conditional representation for decoder preserve much linguistic knowledge (See Section~\ref{subsec:linguisitc}).


\section{Self-Supervised Input Representation}

In this section, we first review the sequence-to-sequence learning in Section~\ref{subsec:preliminaries}. Then we introduce the smoothed data augmentation technique, namely \textsc{Smooth Augmented Data Generator} in Section~\ref{subsec:smooth}. Finally Section~\ref{subsec:BLISS} elaborates our proposed \textsc{Self-Supervised Input Representation} approach.

\subsection{Preliminaries}
\label{subsec:preliminaries}

\paragraph{Sequence-to-Sequence Learning}
Given the target sequence $\bold{y}=\{y_1,y_2,...,y_t\}$ conditioned on a source sequence $\bold{x}=\{x_1,x_2\,...,x_s\}$, the objective of Seq2Seq model can be formulated as $\bold{\hat{y}}=argmax\ logP(\bold{y}|\bold{x})$. Up to now, Transformer~\cite{Vaswani:17} advanced the state of art results compared to other architectures. 
Thus, we employ Transformer as the strong baseline and the test bed.
Transformer consists of an encoder equipped with several identical layers mapping the source sequence $\bold{x}$ into intermediate representation $\bold{h}$ and a decoder equipped with several identical layers taking  $\bold{h}$ and shifted target sequence as input and generates $\bold{y}$ target sequences autoregressively:
\begin{align*}
\bf h &= enc(\bf x+ pos\_emb(\bf x))  \\
\bf {\hat y}_{\leq t} &= dec(\bf {\hat y}_{<t}+ pos\_emb(\bold {\hat y}_{<t}), \bf h)
\end{align*}
Encoder and decoder consists of position-wise feed-forward network, multi-head dot-product attention network and so on, details can be seen from Figure~\ref{fig:transformer}. Noticeable, tokens and position embeddings calculated by $pos\_emb(\cdot)$ are taken as one of the inputs of encoder or decoder, which provides supports for the existence and necessity of position information. So, we design position auxiliary task to restore position information in encoder representation. Decoder generate target sentence autoregressively until meeting special symbol $<eos>$. Finally, the output of the decoder $\hat{\bold{y}}$ is projected into the probability $P(\bold{y}|\bold{x})$, and the optimization objective can be formulated as:
\begin{equation}
 \text{argmax}_{\theta} \text{log}P(\bold{y}|\bold{x};\theta_{enc}, \theta_{dec}) \label{eq:nll_loss}    
\end{equation}
where $\theta_{enc}$ and $\theta_{dec}$ denote the parameters of the encoder and decoder respectively.

\subsection{Smooth Augmented Data Generator}
\label{subsec:smooth}

As shown in Figure~\ref{fig:smooth}, our smooth augmented data generator contains two parts, perturbing functions and smoothness controller.

\paragraph{Perturbing Functions}
As shown in Figure~\ref{fig:smooth}, we feed the source sentences into two perturbing functions, shuffle function and replace function sequentially. For each function, we randomly select $\gamma$ percentage of source sentences for generating augmented data. Specifically, we randomly shuffle tokens within a certain sized window by the shuffle function, and randomly replace several tokens in the source sentences with other words form the source vocabulary by the replace function. 

\begin{figure*}[t] 
    \centering
    \includegraphics[width=.8\textwidth,trim=60 70 60 90,clip]{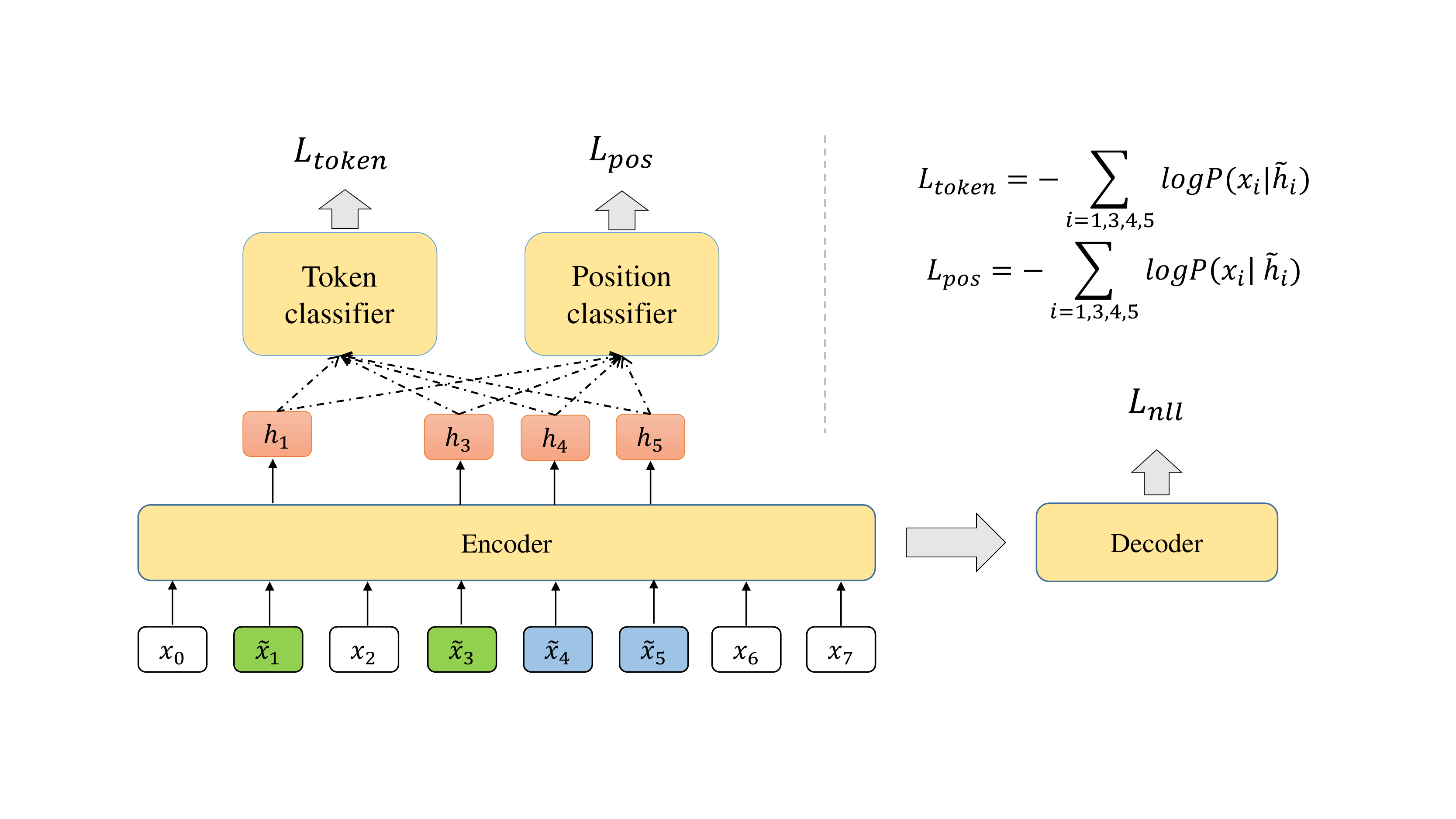}
    \caption{The illustration of our proposed self-supervised input representation (Section~\ref{subsec:BLISS}) in sequence-to-sequence learning framework. We add two classifier to predict the token and position of perturbed tokens synthesized by the smooth augmented data generator in Section~\ref{subsec:smooth}. The meaning of  \textcolor{Liam1}{blue} rectangle and \textcolor{Liam2}{green} rectangle is the same as in Figure~\ref{fig:smooth}. The \textcolor{Liam3}{red} rectangles represent disturbed tokens' intermediate representation produced by the top layer of encoder.}
    \label{fig:supervison}
\end{figure*}
\paragraph{Smoothness Controller}

We set $\alpha_{shu}, \alpha_{rep}$ to control the maximum number of shuffled and replaced tokens respectively. Without smoothness controller, models can only see augmented data with $\alpha_{shu}L$ shuffle tokens of $\alpha_{rep}L$ replaced tokens, where $L$ is the length of sequence. To balance the diversity and similarity of augmented data, we design a smoothness controller to get a smooth distribution of augmented data with different perturbing tokens. Specifically, we hypothesis sampling the number of perturbed tokens $l$ from geometric distribution $l|_{l<=\alpha L} \sim Geometric(p)$, where p is between 0 and 1\footnote{after preliminary studies, we set p=0.2 for all tasks}.  Then, since $l$ is limited by a upper bound, we normalize the distribution of $l$. Finally we sample $l$ 
according to the probability distribution expressed in Equation~\ref{eq:l_sample}. For shuffle function and replacement function, we repeat the above procedures individually and obtain $l_{shu}$ and $l_{rep}$ for perturbing operations. 
\begin{equation}
\label{eq:l_sample}
P(l)= \frac{p(1-p)^{l-1}}{\sum_{i=1}^{\alpha L}p(1-p)^{i-1}}\delta_{1-\alpha L}(l)
\end{equation}

$\delta_{1-\alpha L}(l)$ equal 1 when $l$ is equal or greater than 0 and equal or smaller than $\alpha L$ otherwise 0.

\subsection{Self-Supervised Input Representation}
\label{subsec:BLISS}

Inspired by mask language models~\cite{Devlin:19}, which mask tokens in source sentences and predicts masked tokens on the output, we take similar procedure but two differences distinguish us between them. First, our method is applied in the down-stream tasks with labeled datasets while mask language models take effects in pre-training tasks with unlabeled datasets, so our method works in parallel with mask language models and is complementary with them. Second, prior studies only take token as ground truth label to supervise output. We also take positions as additional labels.

Specifically, we design two softmax classifiers to predict token and position by $\tilde {\bf h}$ respectively. Token classifier are responsible to predict the origin tokens of $\tilde {\bf x}$ while the position classifier predicts the position of perturbed tokens.  And corresponding self-supervision loss functions $L_{token}, L_{pos}$ is expressed  as Equation~\ref{eq:token_loss} and Equation~\ref{eq:pos_loss}, where $x_i, p_i$ denote the origin tokens and absolute position, $W_{token}\in \mathbf{R} ^{e\times v}$ and $W_{pos}\in \mathbf{R}^{e\times p_m}$ represent the parameters of softmax classifier, and $e,v,p_m$ denote embedding dimension, vocabulary size and maximum position index. Following the preliminary trials, we set $p_m=400$. 

\begin{gather}
    \mathcal{L}_{token} = \sum_{i} logP(x_i|\tilde{h}_i,W_{token},\theta_{enc}) \label{eq:token_loss} \\
    \mathcal{L}_{pos} = \sum_{i} logP(p_i|\tilde{h}_i, W_{pos},\theta_{enc}) \label{eq:pos_loss} \\
    \mathcal{L}_{nll}(\bold{\tilde{x}},\bold{y}) = \text{log}P(\bold{y}|\bold{\tilde{x}};\theta_{enc}, \theta_{dec}) \label{eq:nll}
\end{gather}

By integrating the above two loss functions with the traditional negative log-likelihood loss function as Equation~\ref{eq:nll}, the complete objective function of our model is expressed as Equation~\ref{eq:loss}:
\begin{equation}
\label{eq:loss}
\underset{\theta}{\text{argmax}}\ \mathcal{L}(x, y) = \mathcal{L}_{nll} + \lambda_{token}\mathcal{L}_{token}+\lambda_{pos}\mathcal{L}_{pos}
\end{equation}
where $\theta=\{\theta_{enc},\theta_{dec},W_{token},W_{pos}\}$, $\lambda_{token}$ and $\lambda_{pos}$ are hyper-parameters that balance the weights of different self-supervision objectives.

{In conclusion}, we add smooth augmented data generator for source sentences, and restore it's original token and position information on the encoder output. The basic intuition behind is that although the augmented sequence information is  distorted locally, but the contextual information is still preserved, a robust encoder should own the ability to restore correct information from the distorted sentences. Besides the tokens of sequence, the positions of tokens in the sequence play an importance role of the sequence information. So, we design the encoder to predict the positions of swapped tokens to help encoder understand the position information of sequence.

\section{Experiments}
\label{sec:exp}
\begin{table*}[t]
\centering
\caption{Statistics of the datasets and hyperparameters for the experiments. “Batch” denotes the number of
source tokens and target tokens used in each training step. “DP” denotes the dropout value~\cite{Srivastava:14}. “LP” denotes the length penalty~\cite{wu2018pay}. For GEC and text summarization tasks, we chose the checkpoint with best validation score for testing, for translation tasks, we choose the average of last five checkpoints for testing.}
\scalebox{1.2}{
\begin{tabular}{llrrrrrrrr}
\toprule
\multirow{2}{*}{} &\bf Vocab &\multicolumn{3}{c}{\bf Sents} & \multicolumn{3}{c}{\bf Training} & \multicolumn{2}{c}{\bf Testing}   \\ 
&Src/Tgt            &Train &Dev  &Test  &Batch  &Step &DP &Beam &LP \\
\midrule
WMT14 En-De &32768 &4.5M &3K &3K    &64K &300K &0.2     &5 &0.6 \\
\midrule
WMT16 En-Ro &34976 &0.6M &2K &2K    &160K &15K &0.3      &5 &1.0 \\
\midrule
IWSLT14 De-En &10148 &160215 &7282 &6750  &32K &20K &0.3     &5 &1.0 \\
\midrule 
CNN/DM &50264 &0.3M &13K &11K   &64K &70K &0.1  &4 &2.0\\
\midrule 
CONLL &33352 &1.3M &5K &1K &    64K &85K &0.2   &6 &0.6 \\
\bottomrule
\end{tabular}}
\label{tab:datasets}
\end{table*}

\begin{table}[h]
\centering
\caption{Hyper parameters of our methods in different tasks.}
\begin{tabular}{llrrrrr} 
\toprule
\multirow{1}{*}{} &$\gamma$  &$\alpha_{shu}$  &$\alpha_{rep}$    \\ 
\midrule
WMT14 En-De &0.3 &0.1 &0.1  \\
\midrule
WMT16 En-Ro &0.4 &0.1 &0.1 \\
\midrule
IWSLT14 De-En &0.3 &0.12 &0.15   \\
\midrule 
CNN/DM &0.4 &0.08 &0.15 \\
\midrule 
CONLL &0.3 &0.12 &0.1 \\
\bottomrule
\end{tabular}
\label{tab:hyper}
\end{table}

\subsection{Tasks and Datasets}
To validate the effectiveness of BLISS, we conducted experiments on three representative tasks, which vary in the distance between input and output domains and the scale of training data:
\begin{table*}[t]
\caption{Experimental results of the proposed BLISS method on the Seq2Seq tasks. Results marked with $^\dagger$ are statistically significant compared to vanilla transformer, with $^\ddagger$ are statistically significant compared to best baseline.}
\centering
\scalebox{1.2}{
\begin{tabular}{llllllllll} 
\toprule
\multirow{2}{*}{} & \multicolumn{3}{c}{\bf Translation} & \multicolumn{3}{c}{\bf Correction} & \multicolumn{3}{c}{\bf Summarization} \\ 
\cmidrule(lr){2-4} \cmidrule(lr){5-7} \cmidrule(lr){8-10} &\bf De-En &\bf En-Ro &\bf En-De &\bf Prec. &\bf Recall&\bf F$_{0.5}$ &\bf RG-1 &\bf RG-2 &\bf RG-L \\
\midrule
\bf \textsc{Vanilla} &35.1 &34.7 &26.7 &58.7 &33.8 &51.2       &40.1 &17.6 &36.8 \\ 
\hdashline
\bf \textsc{Dropout} &35.4 &35.2 &26.5  &\bf 60.8 &34.2&52.6   &40.4&17.7&37.1 \\
\bf \textsc{Blank}   &35.6 &35.4 &27.0 &59.3 &32.7 &51.0   &40.0&17.5&36.8 \\
\bf \textsc{Shuffle} &34.9 &34.4 &26.5 &52.5 &33.9 &47.3   &40.1&17.3&36.8 \\
\bf \textsc{SeqMix}  &35.4 &\bf 35.5 &26.8 &58.3 &33.5 &50.8       &40.2 &17.6  &36.9  \\ 
\bf\textsc{Switchout}&35.6 &\bf 35.5 &26.9 &60.3 &34.1 &52.3       &\bf 40.6 &\bf 17.9 &37.1   \\ 
\midrule 
{\bf BLISS} (ours)&\bf 35.7$^{\dagger}$ &\bf 35.5$^{\dagger}$ &\bf 27.3$^{\dagger\ddagger}$  & 60.2$^{\dagger}$ &\bf 36.3$^{\dagger\ddagger}$ &\bf 53.2$^{\dagger\ddagger}$ &\bf 40.6$^{\dagger}$ &\bf 17.9$^{\dagger}$ &\bf 37.3$^{\dagger}$ \\
\bottomrule
\end{tabular}}
\label{tab:main}
\end{table*}

\textbf{Machine Translation} takes a sentence in one language as input, and outputs a semantically-equivalent sentence in another language. We evaluate our method on three widely-used benchmarks: 
IWSLT14 German$\rightarrow$English (IWSLT14 De-En\footnote{https://wit3.fbk.eu/},~\cite{Nguyen:19}), WMT16 English$\rightarrow$Romanian (WMT16 En-Ro\footnote{https://www.statmt.org/wmt16/translation-task},~\cite{gu2018non}), and WMT14 English-German (WMT14 En-De\footnote{new://www.statmt.org/wmt14/translation-task},~\cite{Vaswani:17}).
We strictly follow the dataset configurations of previous works for a fair comparison.
For the IWSLT14 De-En task, we train the model on its training set with 160K training samples and evaluate on its test set. For the WMT14 En-De task, we train the model on the training set with 4.5M training samples, where newstest2013 and newstest2014 are used as the validation and test set respectively. As for the WMT16 En-Ro task which has 610K training pairs, we utilize newsdev2016 and newstest2016 as the validation and test set. For each dataset, we tokenize the sentences by Moses~\cite{Koehn:07} and segment each word into subwords using Byte-Pair Encoding (BPE,~\cite{Sennrich:16b}), resulting in a 32K vocabulary shared by source and target languages. All the translation tasks are evaluated with sacreBLEU~\cite{Post:18} score.

\textbf{Grammatical Error Correction} takes a sentence with grammatical errors as input and generates a corrected sentence. We evaluate our method on CONLL14 dataset\footnote{\url{https://www.comp.nus.edu.sg/~nlp/conll14st.html}}, which has 1.4M training samples. We closely follow~\cite{Chollampatt:18} to preprocess the data. The MaxMatch ($\text{M}^2$) scores~\cite{Dahlmeier:12} were used for evaluation with Precision, Recall, and $F_{0.5}$ values.

\textbf{Text Summarization} takes a long-text document as input, and generates a short and adequate summary in the same language. We evaluate our method on the the most representative summarization benchmark CNN/Daily Mail corpus\footnote{\url{https://huggingface.co/datasets/cnn_dailymail}}, which contains 300K training samples. We follow~\cite{Ott:19} to preprocess the data. During testing, the minimum length was set to 55 and the maximum length was set to 140, which were tuned on the development data. We also follow~\cite{Paulus:18} to disallow repeating the same trigram. We evaluate the summarization performance with the standard ROUGE metric~\cite{Lin:04}, i.e. Rouge-1, Rouge-2, and Rouge-L.

The machine translation task has distant input/output domains (i.e. in different languages), while the other tasks has similar input/output domains (i.e. in the same language). Details of the datasets\footnote{We use the same datasets with \cite{liu2021understanding}.} are listed in Table~\ref{tab:datasets}.

\subsection{Implementation}
Our model is based on the Transformer~\cite{Vaswani:17} sequence-to-sequence architecture due to its state-of-the-art performance and all the models are implemented by the open-source toolkit fairseq\footnote{\url{https://github.com/pytorch/fairseq}}~\cite{Ott:19}.
For better reproduction, we employ the base Transformer ($\text{d}_{model}$ = $\text{d}_{hidden}$ = 512, $\text{n}_{layer}$ = 6, $\text{n}_{head}$ = 8) for all tasks in this paper. 
All models were trained on \texttt{NVIDIA DGX A100} cluster.
Table \ref{tab:datasets} gives more details of the benchmarks. It is noted that other unmentioned hyperparameters keep the same with the original paper of Transformer~\cite{Vaswani:17}.
The hyper-parameters of training of different tasks and datasets are listed in Table~\ref{tab:datasets}. 
We set $\lambda_{token}=\lambda_{pos}$=0.005 and $p=0.2$ for all tasks, other hyper parameters varying in tasks as shown in Table~\ref{tab:hyper}. 
 
\subsection{Baselines}

To validate the effectiveness of our methods, we compare our approach with following baselines:

\begin{itemize}
  \item \textsc{\bf Vanilla}~\cite{Vaswani:17}: The original sequence-to-sequence training strategy without any data augmentation strategies.
  
  \item \textsc{\bf Dropout}~\cite{Iyyer:15,Lample:18}: Randomly dropping tokens with their best drop ratio 0.1.

  \item \textsc{\bf Blank}~\cite{Xie:17}: Randomly replacing word tokens with a placeholder, we leave their best setting ratio=0.1 as default.

  \item \textsc{\bf Shuffle}~\cite{Artetxe:18,Lample:18}: Randomly swapping words in nearby positions within a window size K=3.

 \item \textsc{\bf SeqMix}~\cite{Guo:20}: Mixing sentence pairs on both the source and target side. 
 We reimplement according to their public code\footnote{\url{https://github.com/dguo98/SeqMix/tree/main}}.

 \item \textsc{\bf SwithOut}~\cite{Wang:18}: Replacing tokens with other tokens in vocabulary on the source side and target side. We re-implement according to the Appendix A.5 of their paper.

\end{itemize}

\begin{table*}[t]
\caption{Effects of removing each component. The metrics and datasets are same as that of Table~\ref{tab:main}. Bold represents the settings with the most performance degradation for each corresponding task.}
\centering
\scalebox{1.2}{
\begin{tabular}{lrrrrrrrrr} 
\toprule
\multirow{2}{*}{} & \multicolumn{3}{c}{\bf Translation} & \multicolumn{3}{c}{\bf \bf Correction}   & \multicolumn{3}{c}{\bf Summarization} \\ 
\cmidrule(lr){2-4} \cmidrule(lr){5-7} \cmidrule(lr){8-10} &\bf De-En  &\bf En-Ro &\bf En-De  &\bf Prec. &\bf Recall&\bf F$_{0.5}$ &\bf RG-1 &\bf RG-2 &\bf RG-L\\
\midrule
\bf Vanilla     &35.1  &34.7  &26.7   &58.7 &33.8 &51.2       &40.1 &17.6 &36.8 \\ 
 
\midrule 
{\bf BLISS} &35.7 &35.5 &27.3         &60.2&36.3&53.2         &40.6 &17.9 &37.3 \\
\hdashline
 \ -aug-smooth &\bf 35.5 &\bf 35.1 &\bf 26.8        &60.8 &32.3&52.3          &\bf 40.4 &\bf 17.6 &\bf 36.8  \\
 \ -smooth &35.6 &35.2 &26.9       &\bf 60.2&34.9&52.6        &\bf 40.4 &17.7 & 37.0  \\
  \ -token  & 35.6 &\bf 35.1 &\bf 26.8       &60.5&\bf 32.0&\bf 51.3         &\bf 40.4 &17.7 & 37.0  \\
 \ -pos  &35.6  &35.3 &26.9          &60.3&35.0&52.7         &40.5 &17.8 &37.1 \\
\bottomrule
\end{tabular}}
\label{tab:ablation}
\end{table*}
 

\subsection{Main Results}
\label{subsec:mainresults}
Table~\ref{tab:main} lists the performances of our models as well as strong baseline models on different tasks. Clearly, the proposed self-supervised input representation approach (``BLISS'') significantly outperforms the vanilla Transformer in all cases, while there are still considerable differences among model variations. 
Specifically, on translation task, our BLISS equipped models achieve the best among all contrasted approaches, and encouragingly outperform the vanilla transformer by averaged \textbf{+0.7} BLEU points.
As for the grammatical error correction task, we achieve the \textbf{+2.0} F$_{0.5}$ scores improvement against the vanilla model, and notably, our robust self-supervised input representation approach recalls significantly more potential grammatical errors, i.e. \textbf{+2.5} percent. On the contrary, the existing data augmentation approaches, e.g. Shuffle, Blank and SeqMix, slightly undermine the GEC performance. We conjecture that such performance degradation for previous approaches is due to the lack of generalization across tasks, i.e. they are proposed for MT.
As for summarization task, the results also show a promising trend against all baseline methods.
All those findings demonstrate that our proposed robust self-supervised input representation approach (``BLISS'') is effective and universal across language pairs and tasks.

\subsection{Analysis}
\label{subsec:analysis}
In this section, we provide some insights into when and why our BLISS works.

\paragraph{\bf Effects of Each Component}
There are four carefully designed components:
\romannumeral1) \textbf{perturbing functions} named as ``aug'' performs shuffling and replacement operations to generate augmented data sequentially. 
\romannumeral2) \textbf{Smoothness controller} named as ``smooth'' generates augmented data of various degrees.
\romannumeral3) \textbf{Token self-supervision loss} named as ``token'' supervises the lexical information of augmented input, which helps the encoder capture robust token representation.
\romannumeral4) \textbf{Position self-supervision loss} named as ``pos'' supervises the position information of augmented input. 
To verify the reasonableness of those components, we remove different components in Table~\ref{tab:ablation}, e.g. ``-aug-smooth'', ``-smooth'', ``-token'' and ``-pos'', as the ablation settings.
{\bf Takeaway:}
\textit{Our proposed BLISS performs worse when either component is removed, demonstrating the effectiveness of four proposed components.}

\paragraph{\bf Complementary to Related Works}

\begin{table}[t]
\caption{Complementary to other work, i.e. SwitchOut~\cite{Wang:18}. BLISS with SwitchOut-style augmentation function is denoted as ``BLISS w/ S.''. Translation, Correction and Summarization are evaluated with BLEU, F$_{0.5}$ and RG-L, respectively.}
\centering
\scalebox{1.1}{
\begin{tabular}{lccc} 
\toprule
    &\bf WMT14 En-De&\bf CoNLL14 &\bf CNN/DM \\
 \midrule
\bf  SwitchOut & 26.9 & 52.3 & 37.1 \\
\bf BLISS w/ S. & \bf 27.1 & \bf 52.9 & \bf 37.3\\
\bottomrule
\end{tabular}}

\label{tab:complement}
\end{table}

Our proposed BLISS enables self-supervisions from the structure-level, thus BLISS has the great potential to complement existing strong data-level methods. Here we choose SwitchOut~\cite{Wang:18} due to its competitive performance in main experiments. 
We replace the vanilla simple augmentation function in BLISS, i.e. shuffle and replacement, with SwitchOut and the results is reported in Table~\ref{tab:complement}. 
{\bf Takeaway:}
\textit{Our proposed structure-level self-supervised approach BLISS achieves further improvement across different sequence-to-sequence tasks with advanced data augmentation functions, e.g. SwitchOut, showing its appealing expandability.} 

\paragraph{\bf BLISS is Robust to the Inference Noises}
\begin{figure}[t] 
    \centering
    \includegraphics[width=0.5\textwidth,trim=0 3 0 3,clip]{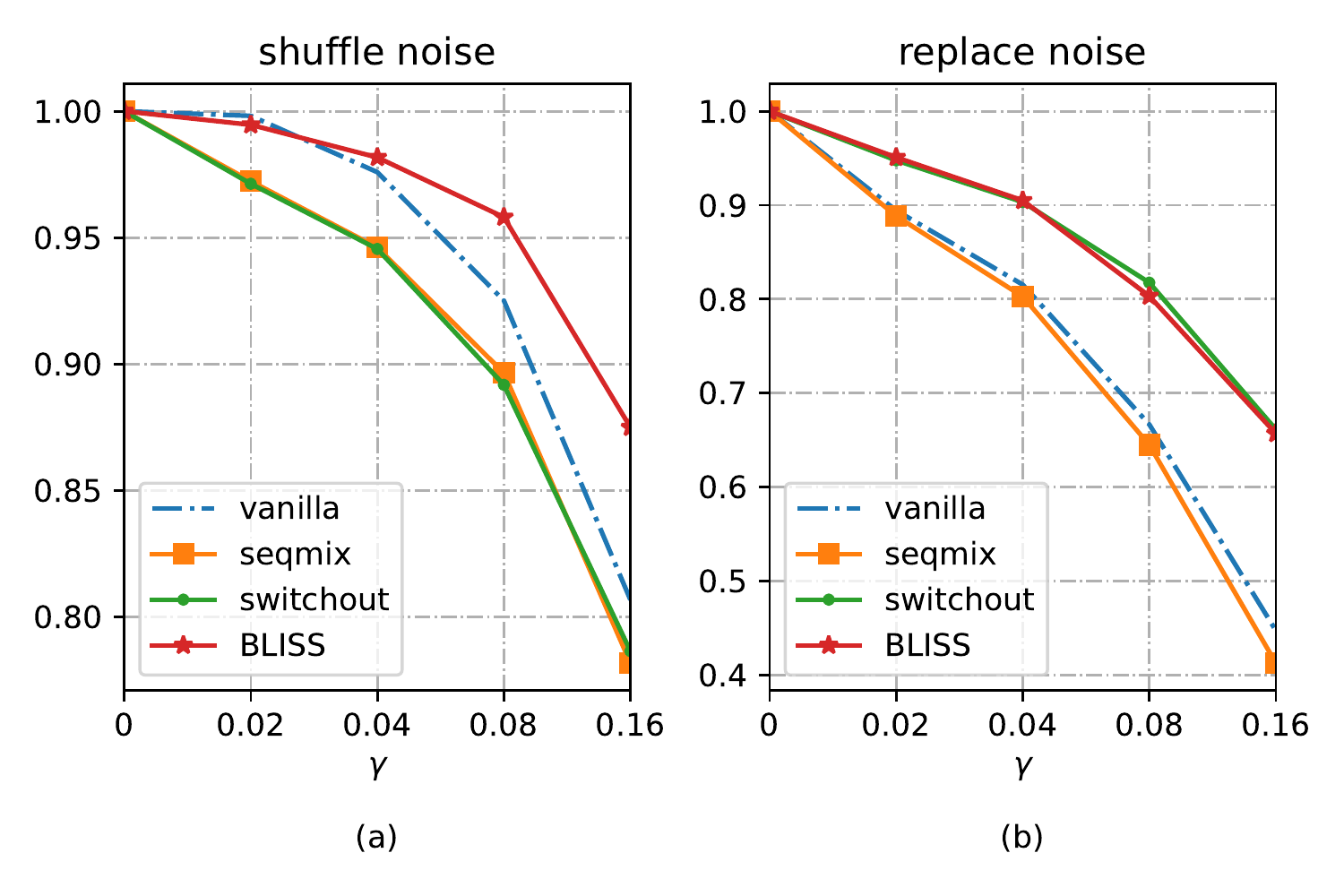}
    \caption{The performance drop when inference on noised testing data, where we test on WMT14 En-De and report the scaled BLEU scores. The noise types for the left and right figures are shuffling and replacing, respectively.} 
    \label{fig:noise}
\end{figure}

Our self-supervised input representation is expected to tolerate the inputting noises to some extent. To validate our hypothesis, we follow \cite{ding-etal-2020-self,zhong2022knowledge} to inject two types of artificial noise, e.g. shuffle and replacement, into the test samples with different ratios ranging from \{2\%, 4\%, 8\% and 16\%\}.
For shuffle noise, we select a span whose length is $\alpha l$ ($l$ is the length of source sentence) and  shuffle the order of words within the span. 
As for the replacement noise, we follow our replacement function, where we randomly replace $\alpha l$ tokens with other tokens in the vocabulary.
Figure~\ref{fig:noise} shows different models' performance on noisy data about WMT14 En-De task. 
{\bf Takeaway:}
\textit{Compared with vanilla Transformer and existing contrastive variants, as noise increases, our model ``BLISS'' is significantly robust to both noise, demonstrating the robustness of our approach.} 

\paragraph{\bf BLISS is Robust to the Hyper-Parameters}

\begin{figure}[t] 
    \centering
    \includegraphics[width=0.5\textwidth,trim=0 6 0 6,clip]{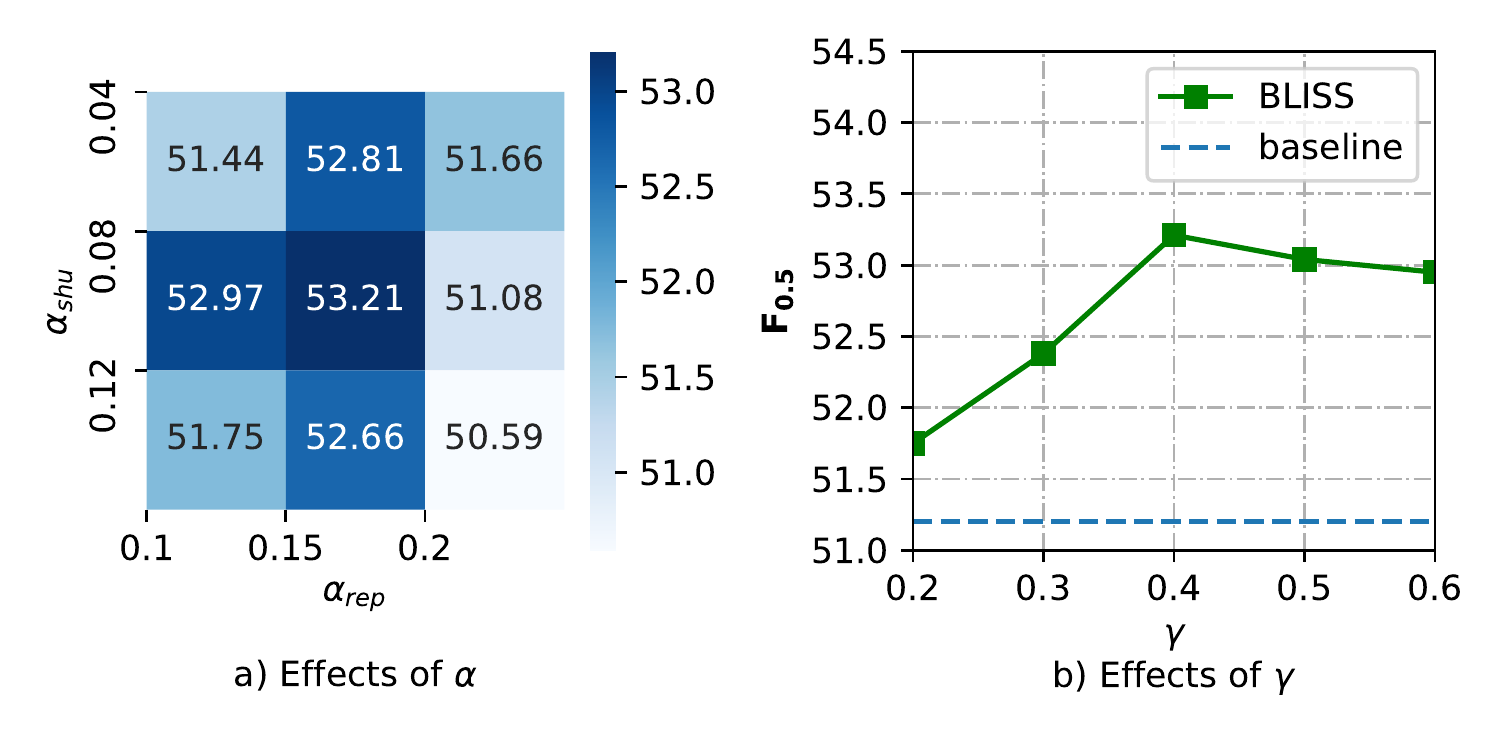}
    \caption{$F_{0.5}$ scores on CONLL14 dataset with different hyper parameters. Left: $\alpha_{shu}, \alpha_{rep}$. Right: $\gamma$.} 
    \label{fig:sens}
\end{figure}

\begin{figure}[t] 
    \centering
    \includegraphics[width=0.4\textwidth,trim=0 0 0 0,clip]{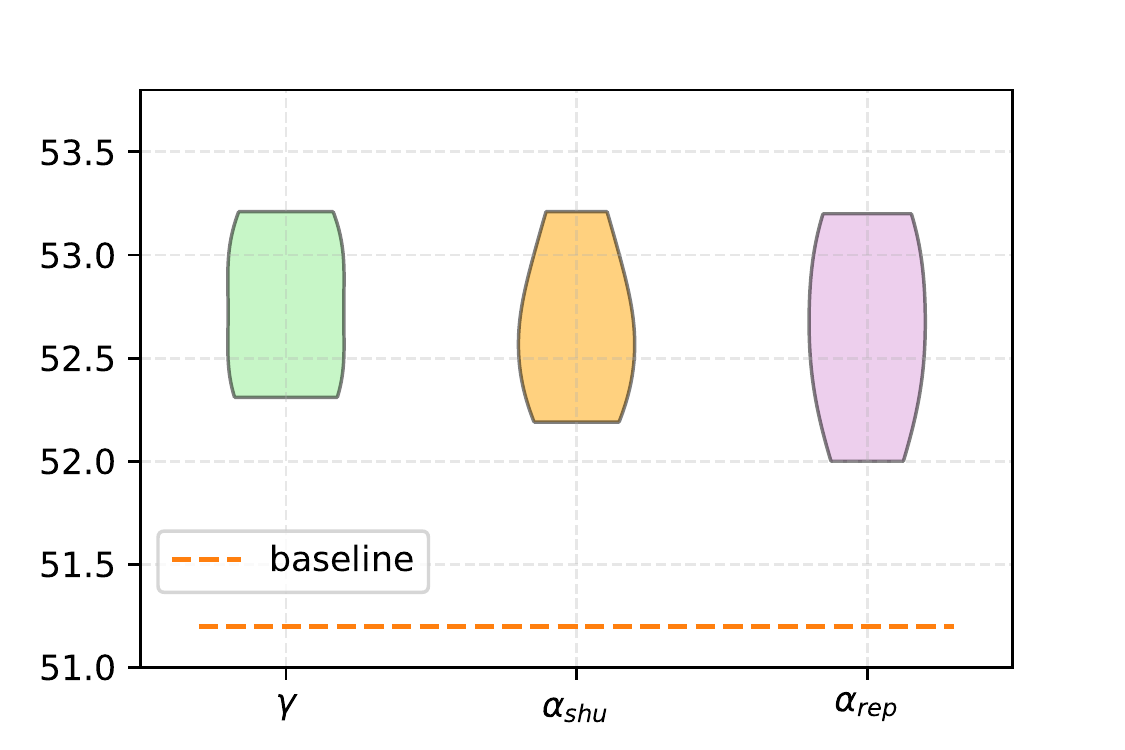}
    \caption{The violin plot of three hyper parameters, $\gamma, \alpha_{shu}, \alpha_{hep}$. The `violin' of each hyper show the distribution of GEC $F_{0.5}$ scores with different value of hyper parameters.} 
    \label{fig:hy-sensitive}
\end{figure}

Data augmentation approaches are always sensitive to hyper-parameters. To dispel the doubt, we investigate whether our approach is robust to different hyper-parameters. We empirically study the effect of hyper parameters $\alpha_{shu}, \alpha_{rep}, \gamma$ on GEC task. We can observe from Figure~\ref{fig:sens} that although the performance varies with hyper-parameters, the extreme values of the results are not significant, still outperforming the baseline approach. To further validate that our model is not hyper-parameter sensitive, we do experiments with different values of hyper-parameters sampling from half of the optimal value to 1.5 times the optimal value. For example, the optimal value of $\gamma$ is 0.04, so we test on values 0.02, 0.03, 0.04, 0.05, 0.06. The violin plot graph is shown in Figure~\ref{fig:hy-sensitive}, where the minimum values of each hyper-parameters are higher than baseline, proving the insensitivity of our hyper-parameters. 

{\bf Takeaway:}
\textit{Our proposed BLISS is not sensitive to hyper-parameters, all hyper-parameters' variants outperform the baseline.}

\begin{table}[t]
\caption{Performance on 10 probing tasks to evaluate the linguistic properties. Note that we train the model on WMT14 En-De.}
\centering
\scalebox{1.2}{\begin{tabular}{llrrr} 
\toprule
\multicolumn{2}{c}{\bf Task} & \bf vanilla & \bf BLISS   \\ 
\midrule

\multirow{2}{*}{\bf Surface} & SeLen &93.1 &\bf 94.0  \\
                             & WC &\bf42.7 & 41.9   \\
\midrule 
\multirow{3}{*}{\bf Syntactic} & TrDep &41.7 &\bf 44.0 \\
                            & ToCo &73.5 &\bf 75.3  \\
                            & BShif &69.3 &\bf 71.8   \\
\midrule 
\multirow{5}{*}{\bf Semantic} & Tense &77.0 &\bf 77.5 \\
                        & SubN &77.3 &\bf 78.4  \\
                        & ObjN &75.0 &\bf 75.2 \\
                        & SoMo & 50.4 &\bf 50.6  \\
                        & CoIn &62.2&\bf 63.3  \\
\bottomrule
\end{tabular}}
\label{tab:probing}
\end{table}

\paragraph{\bf BLISS Captures Better Linguistic Representation}
\label{subsec:linguisitc}
Intuitively, our proposed robust self-supervised input representation approach bringing the capability to correct artificial errors by restoring the token and position information, may help the encoder capture more linguistic knowledge.
To verify this hypothesis, we quantitatively investigate it with 10 probing tasks\footnote{\url{https://github.com/facebookresearch/SentEval/tree/master/data/probing}}~\cite{Conneau:18} to study what linguistic properties are captured by the encoders. 

Each probing task is a classification problem that focuses on simple linguistic properties of sentences.  The 10 probing tasks are categorized into three groups: (1) ``\textbf{Surface}'' focuses on the simple surface properties learned from the sentence embedding. (2) ``\textbf{Syntactic}'' quantifies the syntactic reservation ability; and (3) ``\textbf{Semantic}'' assesses the deeper semantic representation ability. 
More detailed information about the 10 tasks can refer to the original paper~\cite{Conneau:18}.
For each task, we trained the classifier on the train set, and validated the classifier on the validation set. 

Following \cite{Hao:19} and \cite{ding2020context}, we first extract the sentence representations of input sentences by taking average of encoder output. The classifier we used as the sentence as a Multi-Layer Perceptron(MLP) with a hidden dimention of 256. We optimized the model using the Adam optimizer with a leaning rate of 0.001 in 70 epochs for `WC' and `SoMo' task and 10 epochs for other tasks.

To evaluate the representation ability of our BLISS, we compare the pretrained vanilla Transformer~\cite{Vaswani:17} and BLISS equipped machine translation model encoders, followed by a MLP classifier. Sepcifically, the mean of the top encoding layer, as sentence representation, will be passed to the classifier. Table~\ref{tab:probing} lists the results.
{\bf Takeaway:} 
\textit{The proposed BLISS could preserve significant better surface, syntactic and semantic knowledge (Vanilla vs. BLISS = 65.1 vs. 66.2), confirming our hypothesis.} 

\begin{figure}[t] 
    \centering
    \includegraphics[width=0.5\textwidth,trim=0 0 0 0,clip]{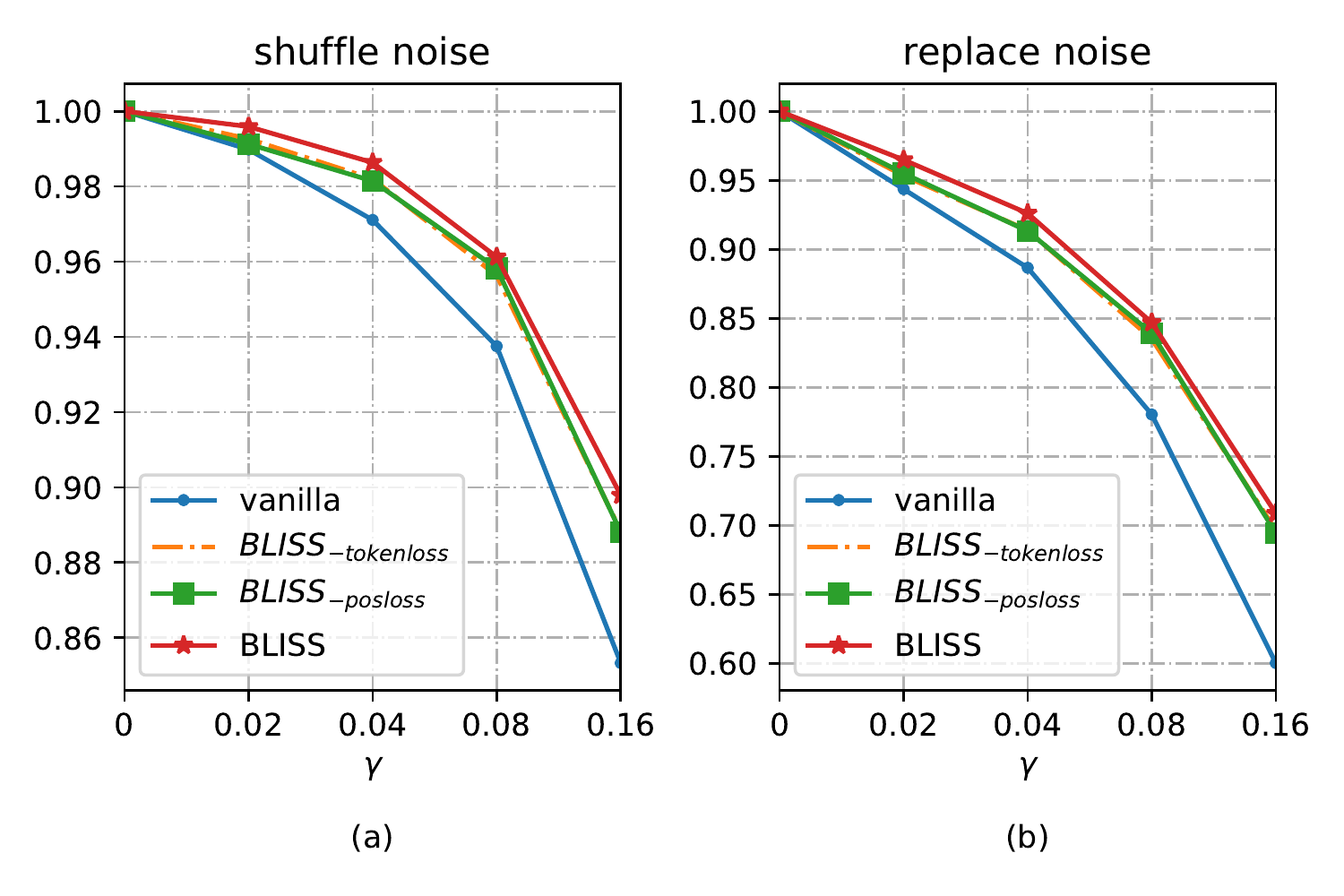}
    \caption{The influence on robustness of models of self-supervision loss functions. We test on IWSLT14 De-En and report the scaled BLUE scores. The x axis $\gamma$ represents the degree of noised data. The noise types for the left and right figures are shuffling and replacing, respectively. Green and orange line represent BLISS models removing position loss and token loss individually.} 
    \label{fig:noise-iwslt}
\end{figure}

\begin{figure}[t] 
    \centering
    \includegraphics[width=0.5\textwidth,trim=0 0 0 0,clip]{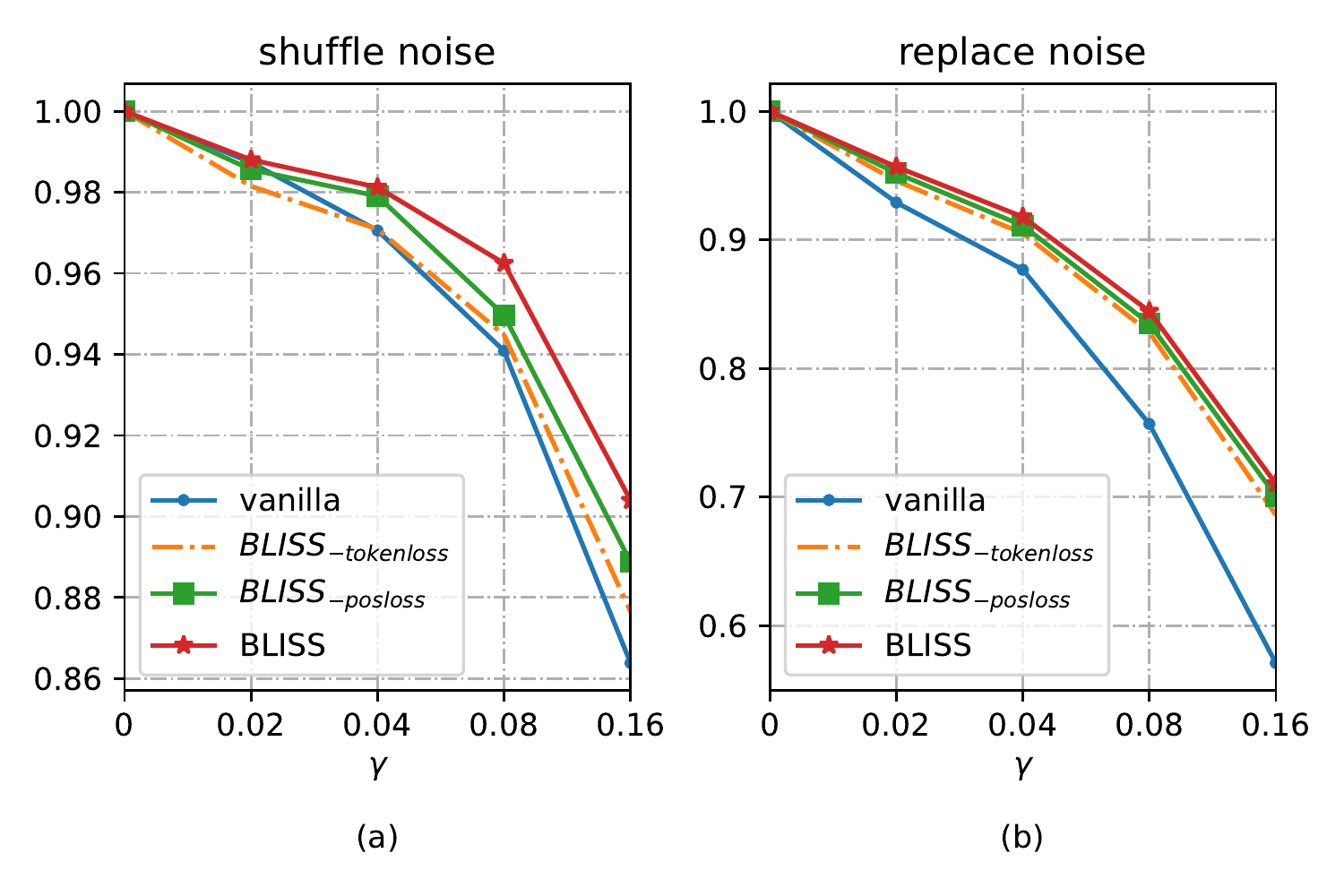}
    \caption{The influence on robustness of models of self-supervision loss functions. We test on WMT16 En-Ro and report the scaled BLUE scores. The meaning of axis is the same as Figure~\ref{fig:noise-iwslt}.} 
    \label{fig:noise-enro}
\end{figure}

\paragraph{\bf Auxiliary Self-Supervision Loss Improves Model Robustness}
\label{sec:loss-robust}

We introduced two auxiliary  self-supervision loss functions (token denoising loss and position denoising loss) to help the encoders learn more robust representation of the source sequences. 
Given the noised input, a vanilla transformer model learn a robust representation from noised data by the joint effort of encoder and decoder, 
while our method encourages the encoder to take more responsibility to not only understand the noised input but also distinguish the noises (with token and position denoising losses).
To illustrate the effects of the auxiliary denoising losses, we conduct ablation studies to observe the performance degradation without token and position self-supervised losses on IWSLT14 De-En (in Figure~\ref{fig:noise-iwslt}) and WMT16 En-Ro (in Figure~\ref{fig:noise-enro}).
As expected, removing the self-supervision loss function will cause significant performance degradation when performing the noise attacks.
However, the improvement of self-supervision loss functions is relatively small on IWSLT14 De-En task and when the noise is replace. 
The potential reason may be that the IWSLT14 De-En task contains fewer sentences and the replace denoising task is relatively easy for the model to handle. So the encoder is not necessary to be enhanced by self supervision method.




\section{Conclusion}
In this paper, we investigate how to achieve robust sequence-to-sequence learning with self-supervised input representation. To achieve it, we propose to make the most of supervised signals and self-supervised signals with our proposed BLISS, which consists of a smooth augmented data generator and corresponding self-supervised objectives upon the top of the encoder. Experiments show that BLISS consistently outperforms the vanilla Transformer and other five data augmentation approaches in several datasets. Additional analyses show that BLISS indeed learns robust input representation and better linguistic information, confirming our hypothesis. 

Future directions include validating our findings on more sequence-to-sequence tasks (e.g. dialogue and speech recognition) and model architectures (e.g. DynamicConv~\cite{wu2018pay}). Also, its worthy to explore our method to large scale sequence-to-sequence language model pretraining (e.g. BART~\cite{lewis2020bart}).

\bibliography{main}
\bibliographystyle{bibtex/IEEEtran.bst}

\end{document}